\documentclass[conference]{IEEEtran}
\usepackage{times}
\usepackage[numbers]{natbib}

\usepackage{multicol}
\usepackage{multirow}
\usepackage{siunitx}
\usepackage{balance}
\usepackage[bookmarks=true]{hyperref}
\ifdefined
\fi
\usepackage{amsmath,amssymb}
\usepackage{graphicx}
\usepackage{xcolor}
\usepackage{url}
\usepackage{comment}
\usepackage{float}
\usepackage{subcaption}
\usepackage{booktabs}   
\usepackage{makecell}   

\begin{document}
\title{Beyond Global Instructions: Stage-Aware Coordination for Long-Horizon VLA Manipulation}

\author{Jingwei Zhang, Xiangrong Sun, Lu Zhang, Zhi-yong Liu}

\IEEEoverridecommandlockouts
\IEEEaftertitletext{%
\vspace{-0.15in}
\begin{center}
\includegraphics[width=0.98\textwidth]{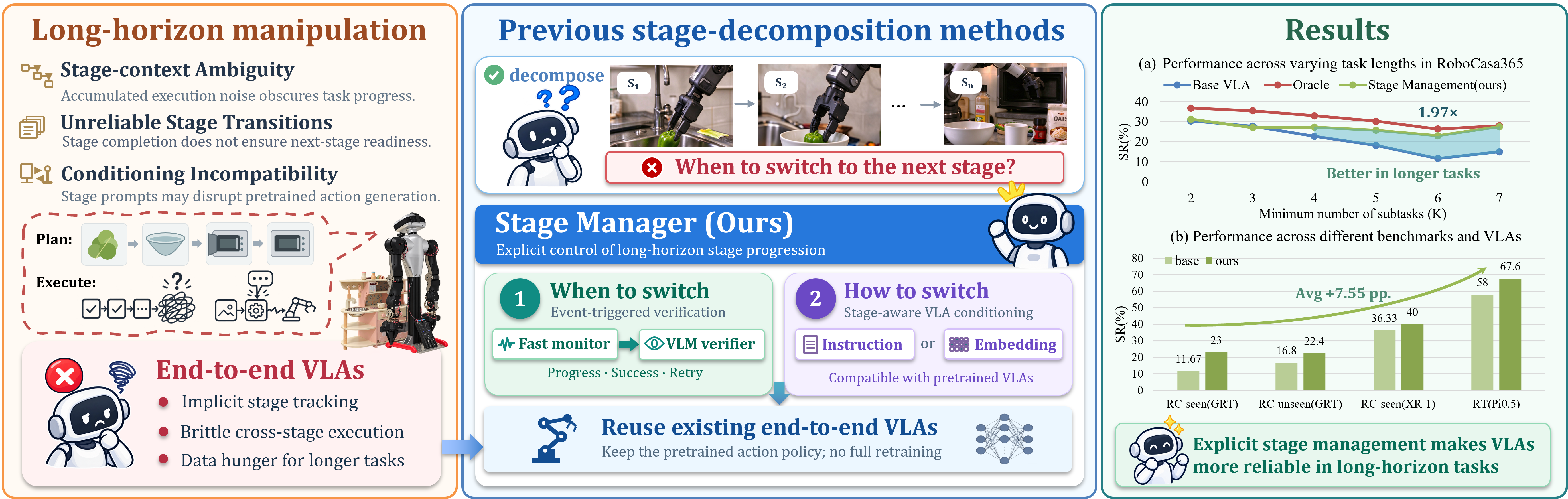}
\end{center}
\refstepcounter{figure}\label{fig:teaser}
\noindent\small Fig. \thefigure: \textbf{Stage management for long-horizon manipulation.}
	Our framework coordinates a frozen VLA through stage-transition
	verification (\emph{when}) and executor-compatible conditioning
	(\emph{how}), serving as a modular booster for long-horizon tasks.
\vspace{0.08in}
}

\maketitle
\thispagestyle{empty}
\pagestyle{empty}

\begin{abstract}
Despite recent progress in VLA models, achieving reliable long-horizon manipulation remains a significant challenge. Our experiments reveal that model success rates drop sharply as task length increases; we hypothesize that as execution unfolds, both task progress and cross-stage dependencies are obscured—issues that current end-to-end approaches have yet to resolve. Based on this observation, we first identify two interrelated bottlenecks in coordinating the different stages of a task: (1) \emph{When to switch}: verifying both current-stage completion and next-stage readiness without privileged simulator states or frequent VLM queries; (2) \emph{How to switch}: conveying stage information without disrupting the pretrained executor's action generation. To address these issues, we introduce a lightweight stage manager that can be integrated into pretrained VLAs. For \emph{when}, a fast monitor combines execution-status predictions and temporal action discrepancy to invoke VLM-based semantic verification. For \emph{how}, we employ context-preserving prompting or a lightweight stage residual adapter for prompt-sensitive executors. Experiments show our approach achieves increasingly pronounced gains on longer task compositions. On six RoboCasa365 tasks with at least six subtasks, our stage management improves GR00T N1.5 success from 11.67\% to 23.00\%, and XR-1 success from 36.33\% to 40.00\%. On five RoboTwin 2.0 tasks, $\pi_{0.5}$ improves from 58.00\% to 67.60\%. These findings highlight explicit stage coordination and executor-compatible conditioning as important directions for reliable long-horizon VLA manipulation.

\end{abstract}

\IEEEpeerreviewmaketitle
\section{Introduction}

The problem of long-horizon robotic manipulation has long attracted significant attention, as it directly relates to the reliable execution of multi-stage tasks in open-world environments. Earlier approaches to this problem
largely relied on explicit skill libraries along with transition rules that were either predefined or learned~\cite{adversarial_skill_chaining_2021,saycan_2022}, often limiting flexibility and
generalization across diverse environments. In contrast,
recent vision-language-action (VLA) models
have substantially advanced robotic manipulation
through end-to-end learning, enabling more flexible behaviors
across diverse tasks~\cite{openvla_2024,groot_2025,pi05_2025,xiaomi_2026}.
However, our empirical evaluation reveals a striking
limitation: pretrained VLAs exhibit a sharp decline in
success as evaluation shifts toward tasks with more subtasks,
as shown in Fig.~\ref{fig:teaser} Results (a).

We therefore argue that stronger end-to-end action generation alone
does not guarantee robust long-horizon execution.
As task lengths extend, possible intermediate states
and execution branches proliferate, while dependencies
across subtasks become increasingly complex.
Learning to handle these variations within a single
end-to-end policy demands extensive and diverse
demonstrations, making it difficult to cover the full
range of execution outcomes and recover from
out-of-distribution (OOD) deviations.
Moreover, even with larger models and longer context
windows, processing the entire multi-stage instruction
at once may dilute attention to the active stage and
obscure intermediate dependencies, resembling the
\emph{lost-in-the-middle} phenomenon observed in
long-context language models~\cite{liu2024lost}.
Consequently, the current observation and global
instruction may provide insufficient cues for tracking
task progress and resolving cross-stage dependencies,
leading to premature actions, skipped steps, or
semantic confusion.

These challenges motivate incorporating \emph{explicit
	stage management} into pretrained VLAs, decomposing
long-horizon execution into manageable stages while
preserving their generalizable action-generation
capabilities. However, explicit coordination introduces
two coupled challenges. First, \emph{when to switch}
requires verifying not only current-stage completion but
also next-stage readiness (e.g., an opened drawer may
still leave its contents inaccessible for grasping).
This process must avoid oracle information unavailable in real-world settings and minimize VLM calls to preserve real-time performance.
Second, \emph{how to switch} requires conveying the
active stage without disrupting pretrained action
generation. Our experiments reveal that even with
oracle transition timing, incompatible stage conditioning
can substantially degrade execution for certain VLA
backbones. Together, these observations identify
stage-transition verification and executor-compatible
conditioning as two critical bottlenecks in long-horizon
manipulation.

Prior works have explored learned completion signals
\cite{seqvla_2025,cyclevla_2026}, VLM coordination
\cite{stageguard_2026,goal2skill_2026}, and steerable
interfaces~\cite{steerable_policies_2026,harness_vla_2026},
but their joint treatment under frozen VLA backbones
remains limited. We address this gap with a lightweight
\emph{stage manager} that decouples
\emph{when} and \emph{how} decisions. For \emph{when},
an event-triggered fast monitor combines an execution-status
head and temporal action discrepancy to invoke VLM-based
verification of stage completion and next-stage readiness.
For \emph{how}, a two-tier conditioning interface conveys stage information
through context-preserving prompting or a
stage-AdaLN adapter, without updating VLA weights.
Our framework
serves as a modular long-horizon booster that can be integrated
with existing VLA executors.
Our main contributions are:
\begin{itemize}
	\item \textbf{What we observed:}
	We observe a pronounced degradation in pretrained VLA
	performance on increasingly long task compositions,
	highlighting the limitations of implicit stage
	coordination in end-to-end VLAs.
	
	\item \textbf{What we identified and proposed:}
	We identify two coupled challenges---stage-transition
	verification (\emph{when}) and executor-compatible
	conditioning (\emph{how})---and propose a lightweight
	stage manager that addresses both while keeping
	pretrained VLA weights frozen.
	
	\item \textbf{What we validated:}
	Across 3 VLAs and 2 benchmarks, our method
	improves GR00T N1.5 success from 11.67\% to 23.00\%, XR-1 from 36.33\% to
	40.00\%, and $\pi_{0.5}$ from 58.00\% to 67.60\%.
\end{itemize}

\section{Related Work}
\label{sec:related}

\noindent\textbf{Stage-Transition Monitoring (When to Switch).}
Unlike modular skill chaining~\cite{adversarial_skill_chaining_2021,saycan_2022} or data-intensive end-to-end VLAs~\cite{openvla_2024,groot_2025,pi05_2025,xiaomi_2026}, explicit stage management reuses a frozen VLA executor, where termination detection becomes a primary bottleneck often sidestepped via privileged oracles,
as also emphasized in prior work~\cite{hivla_2026,stageguard_2026}. Recent oracle-free methods either learn lightweight completion, progress, or distilled transition monitors~\cite{seqvla_2025,cyclevla_2026,stageguard_2026}, or query high-level VLMs for subtask labeling, reflective recovery, and primitive routing~\cite{goal2skill_2026,agentic_2025,loopvla_2026,harness_vla_2026}. However, standalone learned monitors struggle with cross-stage preconditions, whereas frequent VLM polling incurs high latency. We bridge both paradigms by fusing fast execution-status and action-discrepancy cues to event-trigger VLM verification of both current-stage completion and next-stage readiness.

\noindent\textbf{Policy Steerability (How to Switch).}
Even with accurate transition timing, stage-specific commands can induce severe instruction-distribution shifts in frozen VLAs, an issue largely overlooked by prior transition monitors~\cite{seqvla_2025,cyclevla_2026,loopvla_2026}. Existing steerable policies resolve stage ambiguity by fine-tuning the VLA backbone~\cite{steerable_policies_2026} or training stage- and subgoal-conditioned policies from scratch~\cite{scil_2025,sadp_2026}, demanding substantial data and compute. In contrast, we keep all VLA parameters frozen and ensure executor-compatible guidance through context-preserving prompting or a lightweight AdaLN residual adapter.

\begin{figure*}[t]
	\centering
	\includegraphics[width=0.98\textwidth]{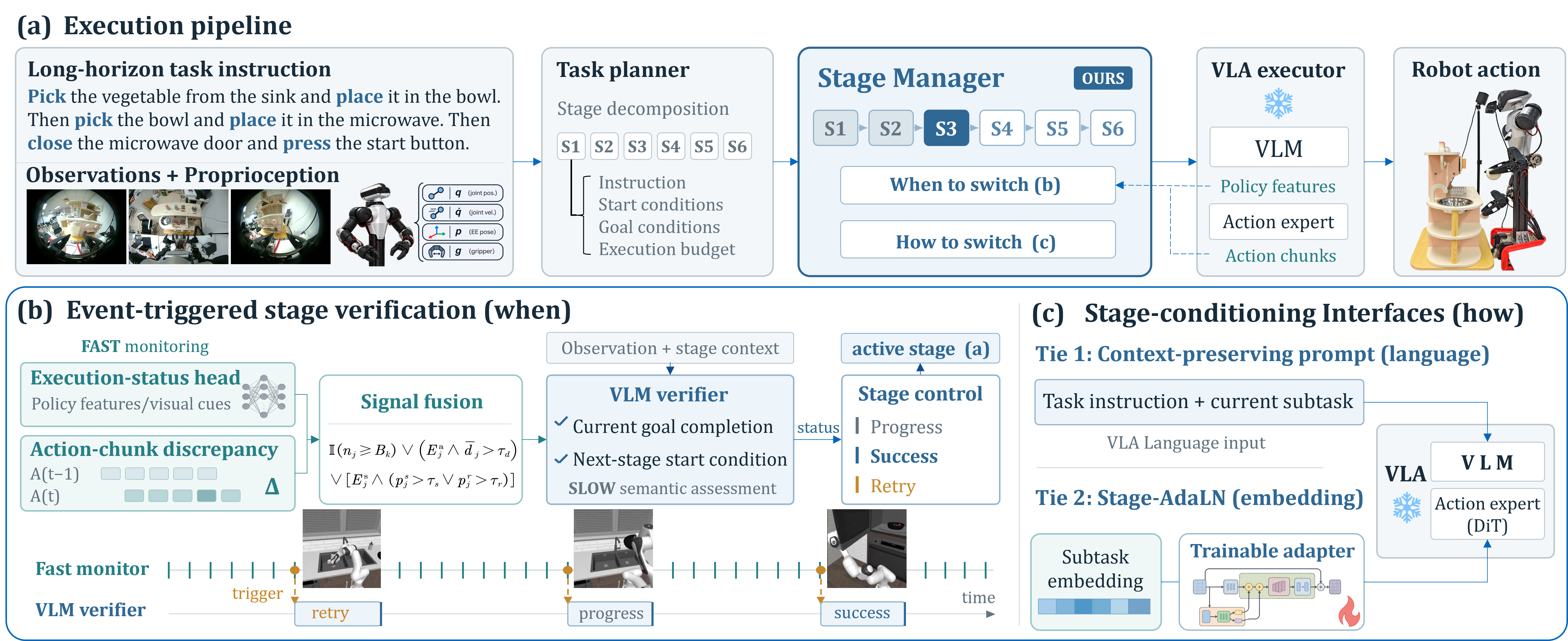}
	\caption{\textbf{Stage-management framework.}
		\emph{When}: A fast monitor triggers VLM-based stage verification.
		\emph{How}: Context-preserving prompting or a trainable AdaLN adapter
		conditions the frozen VLA on the active stage.}
	\label{fig:overview}
\end{figure*}

\section{Stage Management: When and How}
\label{sec:method}

\subsection{Stage Representation and Management}

Given a full-task instruction $\ell$, an LLM planner (e.g. Llama 3.1~\cite{llama3}) generates stage records offline:
$g_k=(\ell_k,s_k^{\mathrm{start}},s_k^{\mathrm{finish}},B_k)$,
where $\ell_k$ describes the stage, $s_k^{\mathrm{start}}$ and
$s_k^{\mathrm{finish}}$ specify its starting and goal conditions
in text, and $B_k$ is the execution budget.
As shown in Fig.~\ref{fig:overview}, the \emph{stage manager}
coordinates between the task planner and
the VLA executor, determining \emph{when} to switch through
event-triggered verification (Sec.~\ref{sec:when}).
Since stage information must guide both transition monitoring
and action generation, we adopt two shared conditioning
interfaces: context-preserving
language prompting (\texttt{LAN}) and feature-level stage
embedding (\texttt{EMB}), detailed in Sec.~\ref{sec:how}.

\subsection{When: Event-Triggered Stage Verification}
\label{sec:when}
We adopt a fast--slow architecture in which an execution-status
head (\texttt{ESH}) estimates stage progress and failures from visual
observations, while temporal action discrepancy (\texttt{TAD}) detects
changes in action predictions. Their event-triggered signals
invoke a VLM to verify current-stage completion and next-stage
readiness, avoiding continuous VLM inference.

\paragraph{Execution-status head (\texttt{ESH})}

A lightweight status head is introduced to
monitor stage progress and failures from visual and
robot-state cues. It supports both conditioning interfaces:
in \texttt{LAN} mode, it reuses cached visual features from the VLA conditioned on the context-preserving instruction (full task + current stage), avoiding additional backbone forward passes during fast monitoring; in \texttt{EMB} mode,
it combines frozen visual tokens with the shared stage
embedding $e_k$, defined in Sec.~\ref{sec:how}, preserving the original
language input. To capture temporal changes, we additionally use
$H=4$ historical visual features. The pooled features,
current state, and historical differences are fed into two
MLPs (4.5M parameters) to predict stage progress
$\hat{r}_j\in[0,1]$ and probabilities $\pi_j^c$ for
\textsc{progress}, \textsc{success}, and \textsc{retry}.

Training uses atomic-task demonstrations and ground-truth
segments from composite-seen demonstrations, with oracle
boundaries used only offline. For each length-$T$ segment,
we assign a linear forward target $r_i=i/(T-1)$ and
synthesize four variations without manual failure annotation:
\textit{reverse} ($r_i=1-i/(T-1)$),
\textit{backtrack} (forward then reversed),
\textit{repeat} ($r_i=r_{i-1}$), and
\textit{mismatch} ($r_i=0$).
The final three forward frames are labeled \textsc{success},
remaining forward frames \textsc{progress}, and reversed,
stalled, or mismatched samples \textsc{retry}.
The \texttt{ESH} is jointly trained with progress regression and
state classification losses, while the VLA remains frozen.

At inference, temporal cues derive an auxiliary state from
sustained high progress
($\min_{0\le m<M}\hat{r}_{j-m}>\tau_{\mathrm{succ}}$)
and progress variation
($\Delta\hat{r}_j=\hat{r}_j-\hat{r}_{j-1}$).
When this state agrees with $\arg\max_c\pi_j^c$, the
corresponding probability is amplified by $\alpha>1$
and renormalized to obtain $p_j^c$; otherwise, the
original probabilities are retained. All tasks share
the same fixed thresholds without task-specific tuning.

\paragraph{Temporal action discrepancy (\texttt{TAD})}

To complement \texttt{ESH}, we measure temporal action discrepancy
between overlapping portions of consecutive action
chunks $A_{j-1},A_j\in\mathbb{R}^{L\times d}$, where $h$
control steps elapse between predictions:
$d_j=\operatorname{RMSE}(A_{j-1}[h:L],A_j[0:L-h])$.
Although \texttt{TAD} cannot distinguish stage transitions from
execution failures, both manifest as large discrepancies
and thus serve as cues for triggering VLM verification. We also apply exponential
moving average (EMA) smoothing:
$\bar d_j=\beta\bar d_{j-1}+(1-\beta)d_j$ to suppress transient fluctuations.

\paragraph{Event-triggered signal fusion}

The fast system combines execution-status predictions and temporal action discrepancy to trigger VLM verification through the following rule:
\begin{equation}
	\begin{split}
		q_j ={}& \mathbb{I}(n_j \ge B_k)
		\lor \left(E_j^{\mathrm{a}} \land \bar d_j>\tau_d\right)\\
		&\lor \left[
		E_j^{\mathrm{s}} \land
		\left(p_j^s>\tau_s \lor p_j^r>\tau_r\right)
		\right],
	\end{split}
	\label{eq:fusion}
\end{equation}
where $q_j$ denotes the VLM trigger decision, $n_j$ is the
steps executed in the current stage, and $\bar d_j$,
$p_j^s$, and $p_j^r$ the smoothed discrepancy and predicted
\textsc{success}/\textsc{retry} probabilities, respectively.
The thresholds $\tau_d$, $\tau_s$, and $\tau_r$, together
with eligibility indicators $E_j^{\mathrm{a}}$ and
$E_j^{\mathrm{s}}$, enforce minimum execution intervals
and post-trigger cooldowns to suppress redundant VLM calls.

\paragraph{VLM-based semantic verification}
When triggered, the VLM (Qwen3-VL-8B-Instruct~\cite{qwen3vl}) receives recent multi-view observations, the full instruction $\ell$, the active stage $\ell_k$, its completion condition $s_k^{\mathrm{finish}}$, and the next-stage precondition $s_{k+1}^{\mathrm{start}}$. It verifies both current-stage completion and next-stage readiness, since completing one stage does not necessarily satisfy the next stage's preconditions. Such task-dependent conditions are difficult to exhaustively specify with hand-crafted rules, motivating VLM-based semantic verification.

The stage manager \textsc{continues} if further
progress is needed and \textsc{advances} when both
current-stage completion and next-stage readiness are
verified. Upon execution failure, the manager
\textsc{retries} by physically backtracking to the
previous verified stage checkpoint (replaying the
recorded actions in reverse order with inverted motion
increments before re-executing the stage).
The selected stage is then conveyed to the VLA through
the conditioning interface in Sec.~\ref{sec:how}.
\begin{figure*}[t]
	\centering
	\includegraphics[width=0.98\textwidth]{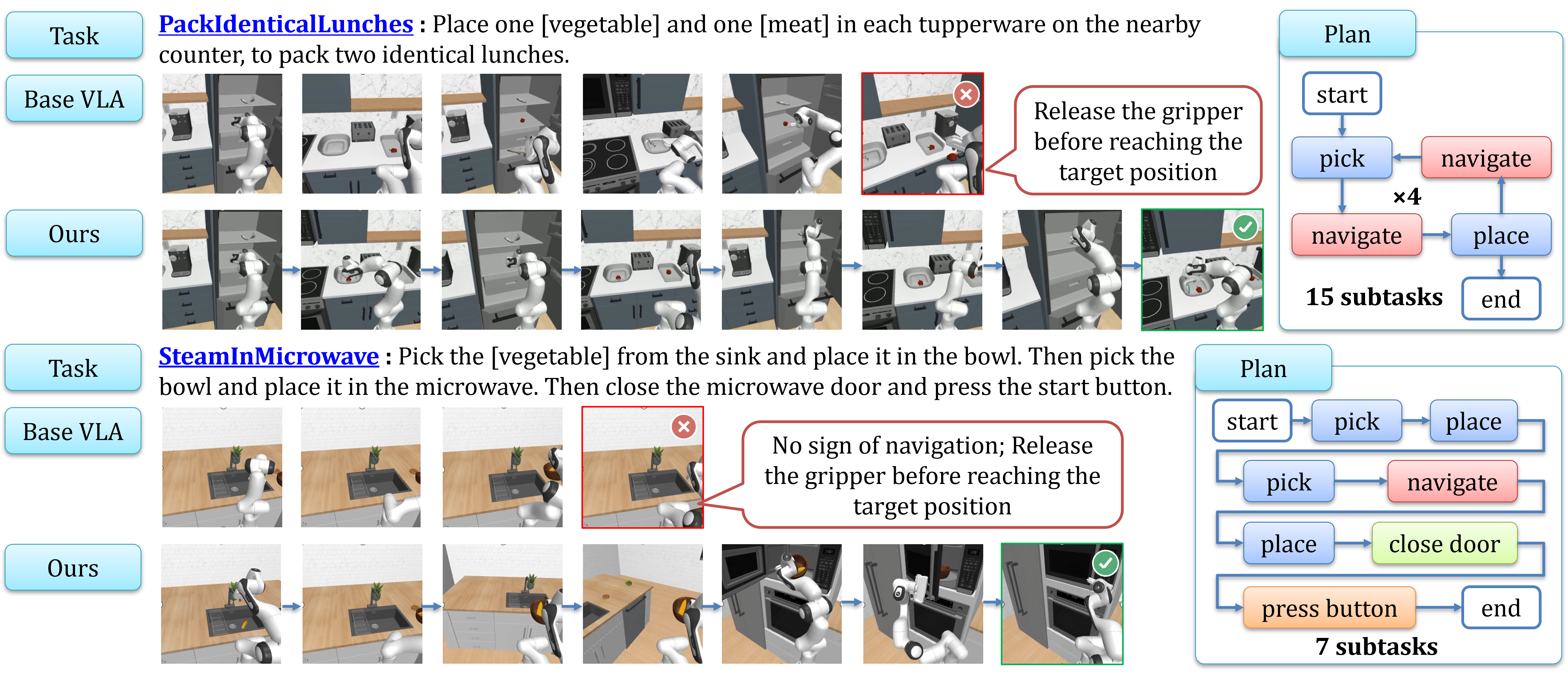}
	\caption{\textbf{Qualitative comparison on long-horizon RoboCasa365 tasks.} As the number of interaction steps increases, the Base VLA suffers from severe global context decay and blurred subtask dependencies. In PackIdenticalLunches (15 subtasks), it prematurely releases the gripper before reaching the target. In SteamInMicrowave (7 subtasks), it completely misses the necessary navigation steps and releases the object prematurely due to accumulated local noise. In contrast, our stage manager explicitly verifies stage transitions and maintains task alignment, successfully executing the complex sequential dependencies without losing track of the global goal.}
	\label{fig:exp}
\end{figure*}
\subsection{How: Stage-Conditioning Interfaces}
\label{sec:how}

Accurate transition timing alone is insufficient if conveying the active stage disrupts the frozen VLA's pretrained conditioning distribution. 
To reconcile zero-shot deployability with executor compatibility, we establish a two-tier conditioning interface governed by a \emph{training-free-first, feature-modulation-fallback} principle. 
By default, we employ training-free context-preserving prompting (\texttt{LAN}). 
If a brief calibration check over a few rollouts reveals prompt-distribution sensitivity---as in XR-1~\cite{xiaomi_2026}, where appending stage text degrades execution even under oracle transitions---we switch to a lightweight, zero-initialized Stage-AdaLN adapter (\texttt{EMB}) that leaves the native language input intact.

\paragraph{Tier 1: Context-preserving prompt (\texttt{LAN})}
For typical VLAs, replacing the original task prompt with an isolated subtask discards global context and induces distribution shift. 
Instead, we retain the full-task instruction $\ell$ and append the active-stage description $\ell_k$ via $\tilde{\ell}_j = \operatorname{concat}(\ell, \ell_k)$, disambiguating the immediate execution goal across long horizons with zero training overhead.

\paragraph{Tier 2: Lightweight Stage-AdaLN (\texttt{EMB})}
For prompt-sensitive VLAs requiring strict invariance in their language pathway, we keep $\ell$ unmodified and inject stage guidance into the action expert's modulation space. 
Leveraging the adaptive layer normalization (AdaLN) pathway standard in DiT policies~\cite{peebles2023dit}, we encode $\ell_k$ through the frozen language backbone as $e_k = \operatorname{Pool}(\operatorname{LM}_{\theta}(\ell_k)) \label{eq:ek}$ and project it via a zero-initialized residual adapter $f_\phi$:
$
	\tilde{b}(\tau, k) = b_\theta(\tau) + \sigma(\gamma) \odot f_\phi(e_k),
	\label{eq:adaln}
$
where $b_\theta(\tau)$ is the pretrained flow-time modulation and $\sigma(\gamma)$ denotes learnable gating coefficients; zero-initializing $f_\phi$ preserves the exact pretrained policy distribution at initialization. 
Keeping VLA weights $\theta$ frozen, we optimize only $\{\phi, \gamma\}$ (5.2M params, $<\!0.2\%$ of a 3B--5B VLA) on stage-segmented demonstrations via flow matching~\cite{lipman2022flow}. 
With $x^\tau = (1-\tau)\epsilon + \tau A^*$ for expert action chunk $A^*$, $\epsilon \sim \mathcal{N}(0, I)$, and $\tau \sim \mathcal{U}(0, 1)$, we minimize
$
	\mathcal{L}_{\mathrm{stage}} = \mathbb{E}\left[ \left\| v_{\theta,\phi,\gamma}(x^\tau, \tau, o, p, \ell, e_k) - (A^* - \epsilon) \right\|_2^2 \right],
	\label{eq:flow}
$
where velocity predictor $v_{\theta,\phi,\gamma}$ conditions on visual observations $o$, robot state $p$, intact global instruction $\ell$, and stage embedding $e_k$, translating subtask semantics into trajectory modulation across broader DiT-based action experts.

\subsection{Long-Horizon Evaluation}

We evaluate our method across tasks, VLAs, and benchmarks. All pretrained VLA parameters remain frozen; only the \texttt{ESH} (and the adapter for tier-2 conditioning) is trained. Each task is evaluated over 50 episodes with fixed seeds.

To examine the effect of task length, we group RoboCasa365~\cite{robocasa365_2026} composite-seen tasks by a minimum subtask threshold \(K_{\min}\), evaluating the average success rate over tasks with \(K \geq K_{\min}\). As shown in Fig.~\ref{fig:teaser} (Results (a)), the performance gap between the base VLA and our stage manager generally widens as shorter tasks are excluded, with the largest relative gain occurring at \(K_{\min}=6\) (1.97×). Although the trend is not strictly monotonic due to varying task compositions, it suggests that explicit stage coordination is particularly beneficial for longer task compositions.
For \(K_{\min}=6\) task subsets, our method improves GR00T N1.5~\cite{groot_2025}
success from 11.67\% to 23.00\%, and XR-1~\cite{xiaomi_2026} from 36.33\% to 40.00\%
(Tab.~\ref{tab:robocasa_seen}).
On 5 composite-unseen tasks with at least 4 subtasks,
success improves from 16.80\% to 22.40\%, though the
remaining gap to oracle (48.4\%) highlights challenges
in generalizing stage management to unseen compositions, it still demonstrates the effectiveness of how-to-switch
(Tab.~\ref{tab:robocasa_composite_unseen}).

\begin{figure*}[t]
	\centering
	\includegraphics[width=0.98\textwidth]{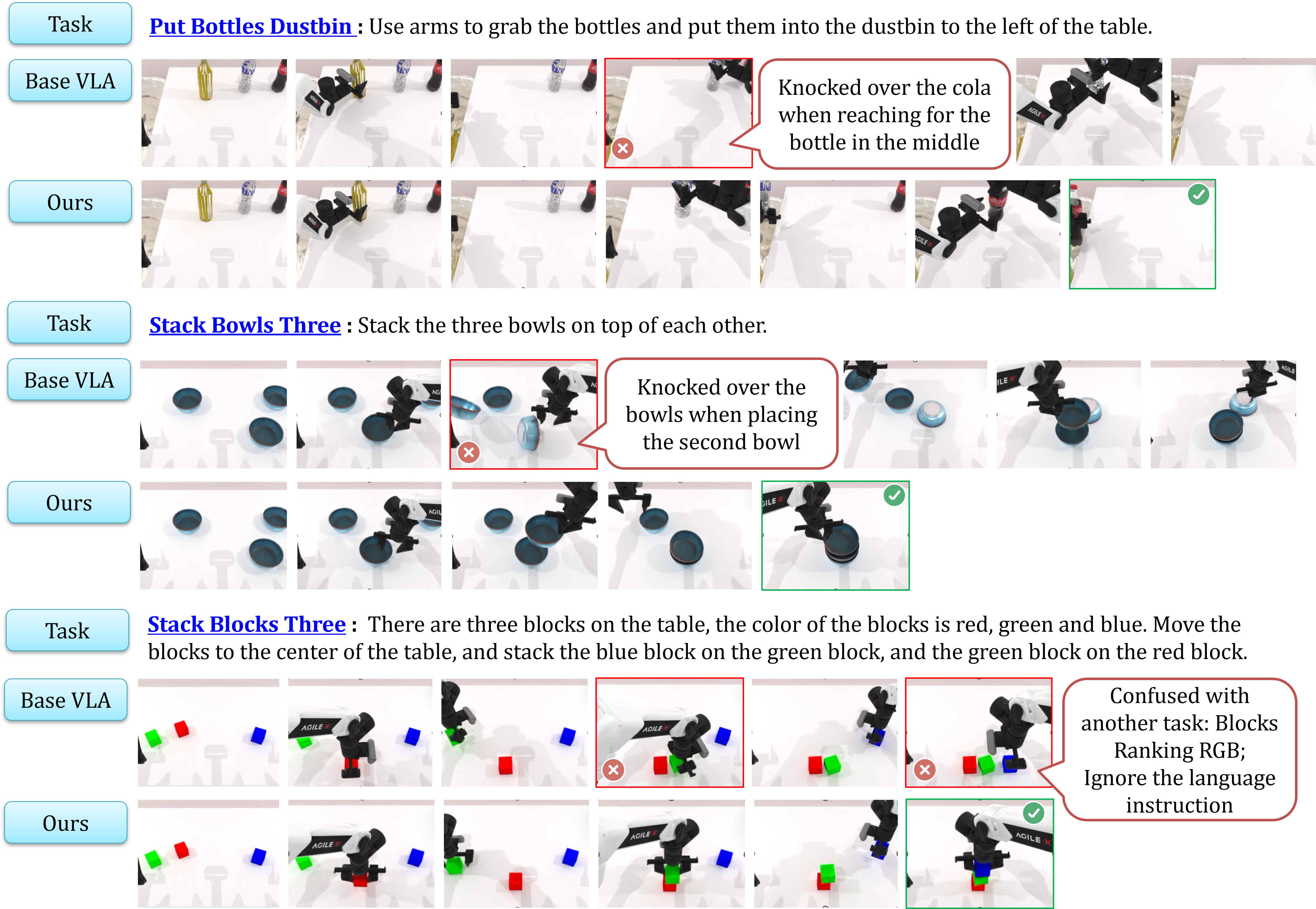}
	\caption{\textbf{Qualitative comparison on long-horizon RoboTwin 2.0 tasks.} We observe that end-to-end VLAs frequently fail when accumulated local physical noise dominates the observation stream, leading to physical collisions (e.g., knocking over the cola in PutBottlesDustbin and overturning bowls in StackBowlsThree) or semantic confusion where the global instruction is ignored (e.g., sorting blocks by RGB instead of the instructed order in StackBlocksThree). Our method effectively mitigates these failures by ensuring stage readiness and re-anchoring the executor to the active stage condition.}
	\label{fig:exp-twin}
\end{figure*}

We further evaluate $\pi_{0.5}$~\cite{pi05_2025}
on 5 RoboTwin 2.0~\cite{chen2025robotwin} tasks.
Our method achieves 67.60\% success, surpassing oracle
transitions (62.00\%), with particularly large gains on
contact-rich tasks such as \textit{Stack Bowls Three}
(84\% vs. 66\%). This suggests that verifying next-stage
readiness can avoid premature transitions caused by rigid
ground-truth boundaries (Tab.~\ref{tab:robotwin}).
Quantitative results are summarized in Tab.~\ref{tab:benchmark}, $K_{\min}$ denotes the minimum number of subtasks among the tasks selected in the benchmark. Qualitative comparisons are shown in Fig.~\ref{fig:exp} and Fig.~\ref{fig:exp-twin}.

\begin{table*}[htbp]
	\centering
	\caption{Per-task success rates (\%) on RoboCasa365 composite-seen
		tasks. Ora denotes oracle stage transitions, with language
		conditioning (LAN) or embedding conditioning (EMB).
		Ours uses online stage management.}
	\label{tab:robocasa_seen}
	\small
	\setlength{\tabcolsep}{5pt}
	\renewcommand{\arraystretch}{1.12}
	\begin{tabular*}{\textwidth}{
			@{\extracolsep{\fill}}
			l c c c c c c c c c
			@{}
		}
		\toprule
		& & \multicolumn{4}{c}{GR00T N1.5 (GRT)~\cite{groot_2025}}
		& \multicolumn{4}{c}{XR-1~\cite{xiaomi_2026}} \\
		\cmidrule(lr){3-6}
		\cmidrule(lr){7-10}
		Task & $K$
		& Base & Ora-\texttt{LAN} & Ora-\texttt{EMB} & Ours-\texttt{LAN} 
		& Base & Ora-\texttt{LAN} & Ora-\texttt{EMB} & Ours-\texttt{EMB} \\
		\midrule
		PackIdenticalLunches  & 15 & 0  & 6  & 6 & 6
		& 20 & 0  & 22 & 22 \\
		SearingMeat           & 6  & 2  & 14 & 8 & 2
		& 2  & 4  & 16 & 12 \\
		SetUpCuttingStation   & 7  & 20 & 40 & 24 & 32
		& 56 & 34 & 64 & 58 \\
		SteamInMicrowave      & 7  & 6  & 34 & 24 & 30
		& 36 & 32 & 46 & 40 \\
		StirVegetables        & 6  & 8  & 32 & 14 & 26
		& 42 & 10 & 42 & 42 \\
		StoreLeftoversInBowl  & 7  & 34 & 32 & 34 & 42
		& 62 & 4  & 70 & 66 \\
		\midrule
		\textbf{Mean} & --
		& 11.67 & \textbf{26.33} $\uparrow$ & 18.33 $\uparrow$ & \textbf{23.00} $\uparrow$
		& 36.33 & 14.00 $\downarrow$ & \textbf{43.33} $\uparrow$ & \textbf{40.00} $\uparrow$ \\
		\bottomrule
	\end{tabular*}
\end{table*}

\begin{table}[htbp]
	\centering
	\caption{SRs(\%) of GRT on RC365 composite-unseen tasks}
	\label{tab:robocasa_composite_unseen}
	\begin{tabular*}{0.95\linewidth}{@{\extracolsep{\fill}} lcccc @{}}
		\toprule
		Task & $K$ & Base & Ora-\texttt{LAN} & Ours \\
		\midrule
		ArrangeTea             & 6 & 32  & 42  & 42 \\
		HeatKebabSandwich      & 5 & 4   & 10  & 4  \\
		PortionHotDogs         & 5 & 6   & 24  & 10 \\
		RecycleBottlesByType   & 4 & 14  & 90  & 26 \\
		WaffleReheat           & 4 & 28  & 76  & 30 \\
		\midrule
		Mean                   & -- & 16.8 & 48.4 & 22.4 \\
		\bottomrule
	\end{tabular*}
\end{table}

\begin{table}[htbp]
	\centering
	\caption{SRs(\%) of $\pi_{0.5}$ on RoboTwin 2.0 tasks}
	\label{tab:robotwin}
	\begin{tabular*}{0.95\linewidth}{@{\extracolsep{\fill}} lcccc @{}}
		\toprule
		Task & $K$ & Base & Ora-\texttt{LAN} & Ours \\
		\midrule
		put\_bottles\_dustbin & 6 & 76 & 86 & 88 \\
		put\_object\_cabinet  & 3 & 40 & 40 & 50 \\
		scan\_object          & 3 & 42 & 42 & 48 \\
		stack\_bowls\_three   & 6 & 66 & 66 & 84 \\
		stack\_blocks\_three  & 6 & 66 & 76 & 68 \\
		\midrule
		Mean                  & -- & 58 & 62 & 67.6 \\
		\bottomrule
	\end{tabular*}
\end{table}
\begin{table}[htbp]
	\centering
	\caption{SRs(\%) across different benchmarks and VLAs.}
	\label{tab:benchmark}
	\begin{tabular*}{0.95\linewidth}{@{\extracolsep{\fill}} llrrrr @{}}
		\toprule
		Benchmark($K_{\min}$) & VLA & Base & Interface & Oracle & Ours \\
		\midrule
		RC-seen(6)   & GRT & 11.67 & \texttt{LAN} & 26.33 & 23.00 \\
		RC-unseen(4) & GRT & 16.80 & \texttt{LAN} & 48.40 & 22.40 \\
		RC-seen(6)   & XR-1 & 36.33 & \texttt{EMB} & 43.33 & 40.00 \\
		RT(3)        & $\pi_{0.5}$ & 58.00 & \texttt{LAN} & 62.00 & 67.60 \\
		\bottomrule
	\end{tabular*}
\end{table}

\begin{figure}[htbp]
	\centering
	\includegraphics[width=\linewidth]{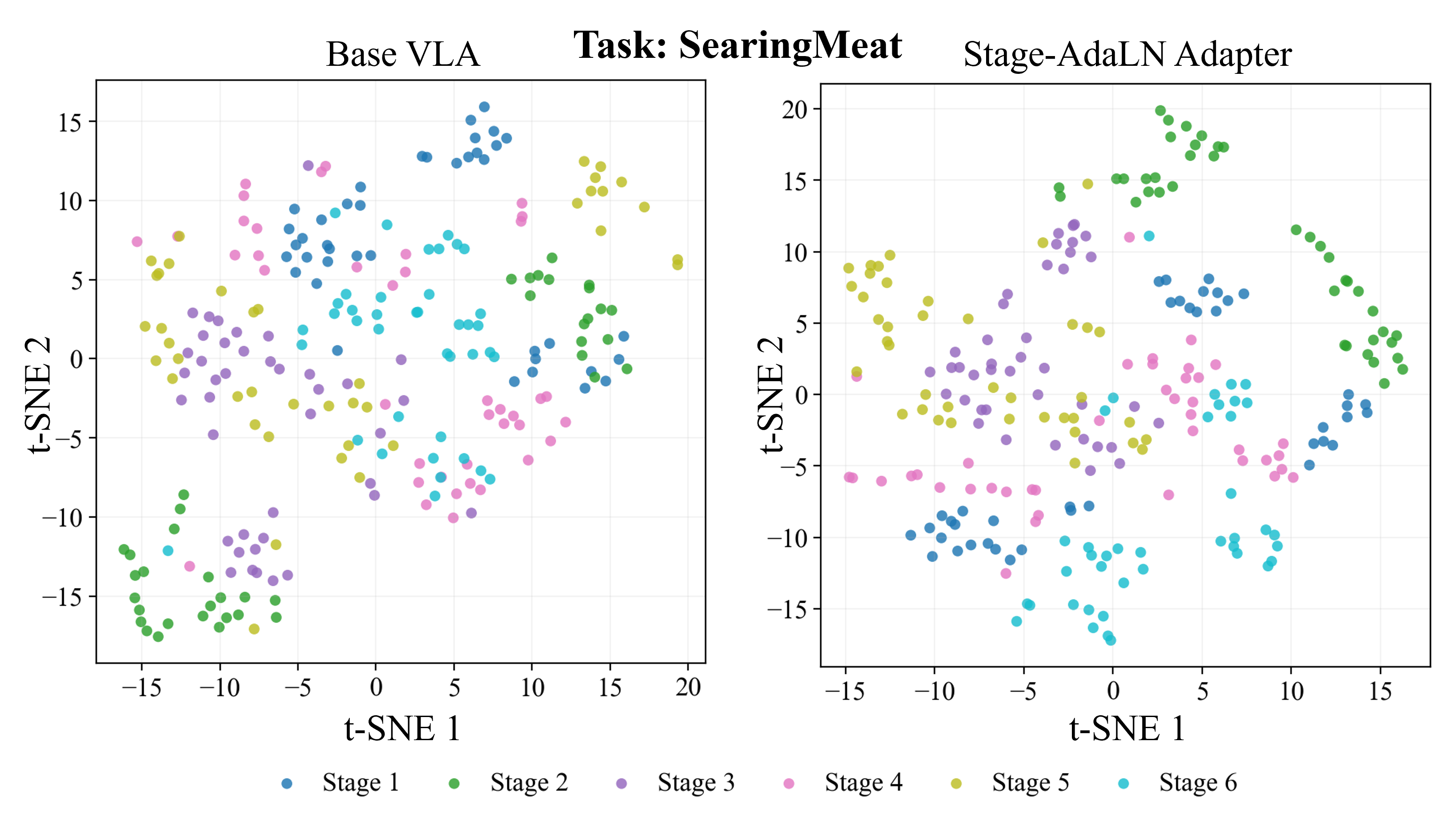}
	\caption{t-SNE visualization of XR-1 DiT features.}
	\label{fig:embedding}
\end{figure}

\subsection{Mechanism Analysis and Ablation Study}

\paragraph{Stage-AdaLN representation analysis}
We analyze the penultimate-layer DiT features of XR-1.
As shown in Fig.~\ref{fig:embedding}, stage embedding
conditioning produces more structured and separable
representations. Across six tasks, stage classification
accuracy improves by 7.42 percentage points on average,
while stage entropy decreases by 0.131, indicating explicit stage conditioning enhances
stage-discriminative representations.

\paragraph{Ablation of stage-transition components (When)}
Tab.~\ref{tab:ablation} shows that \texttt{ESH} and \texttt{TAD} provide
complementary signals: combining both with event-triggered
VLM verification achieves the highest success while reducing
VLM calls by 76.5\% compared with fixed-step verification.
Removing either component or omitting next-stage readiness
verification degrades performance, confirming the importance
of both complementary monitoring cues and readiness-aware
transitions. 

Though the prior methods were not fully replicated, these ablation studies provide controlled comparisons of representative monitoring designs under a shared frozen-VLA backbone. Specifically, \textit{Fixed-step VLM} instantiates periodic polling~\cite{goal2skill_2026,agentic_2025}, while \textit{w/o TAD} reflects learned completion-head triggering~\cite{seqvla_2025,cyclevla_2026}.

\begin{table}[htbp]
	\centering
	\caption{Ablation of stage-transition components.
		calls: Average number of VLM calls per episode;
		$\checkmark\kern-1.2ex\raisebox{1ex}{\rotatebox[origin=c]{125}{\textbf{--}}} $ denotes verifying current-stage completion without next-stage readiness.}
	\label{tab:ablation}
	\small                                   
	\setlength{\tabcolsep}{4pt}              
	\begin{tabular}{@{}lcccccc@{}}           
		\toprule
		Method & \texttt{ESH} & \texttt{TAD}  & VLM & SR (\%) & calls & time (s) \\
		\midrule
		Baseline           & -- & -- & -- & 11.67 & 0     & 89  \\
		Fixed-step VLM     & -- & -- & \checkmark & 19.67 & 35.25 & 166 \\
		w/o \texttt{ESH}            & -- & \checkmark & \checkmark & 17.33 & 10.47 & 117 \\
		w/o \texttt{TAD}            & \checkmark & -- & \checkmark & 16.67 & 9.29 & 114 \\
		Fast only          & \checkmark & \checkmark & -- & 18.33 & 0 & 94 \\
		Current only       & \checkmark & \checkmark & $\checkmark\kern-1.2ex\raisebox{1ex}{\rotatebox[origin=c]{125}{\textbf{--}}} $ & 18.00 & 7.59 & 105 \\ 
		\textbf{Ours}      & \checkmark & \checkmark & \checkmark & \textbf{23.00} & 8.28 & 111 \\
		\bottomrule
	\end{tabular}
\end{table}

\paragraph{Ablation of Stage-Conditioning Interfaces (How)}
To isolate the effect of \emph{how}, we compare
different interfaces under oracle transitions. GR00T N1.5 benefits from
preserving global task context and achieves 26.33\% success, compared with 7.67\%
for subtask-only prompting. The learned adapter is also effective (11.67\% to 18.33\%). In contrast, XR-1 is highly
sensitive to language modifications, with subtask-only
and context-preserving prompting achieving only 14.33\%
and 14.00\%, respectively, while embedding conditioning
achieves 43.33\%. These results support our two-tier design:
context-preserving prompting as the default and Stage-AdaLN
as a lightweight fallback for wider prompt-sensitive backbones.

\section{Conclusion and Future Work}

Our empirical findings reveal that pretrained VLAs
become increasingly unreliable on longer tasks,
suggesting that stronger end-to-end action generation
alone is insufficient for robust long-horizon execution; instead, explicit stage management is required.
We identify two key challenges: determining the timing for switching stages, and conveying stage information without disrupting pretrained executors. Our stage manager further explores several aspects: how to avoid frequent VLM calls, how to integrate evidence from multiple sources to determine execution status, and how to establish appropriate execution interfaces for different VLAs.
We address these challenges while keeping
VLA backbones frozen, achieving consistent improvements
across multiple VLAs and benchmarks, particularly on longer tasks. 
These findings highlight explicit stage coordination
as a promising complement to end-to-end VLAs for
reliable long-horizon manipulation.

While our method demonstrates promising results, it remains an initial exploration with several limitations that point to future directions. First, our evaluation is primarily simulation-based; deploying the stage manager in real-world experiments is essential to validate its robustness against unmodeled physical noise and system dynamics. Second, our current failure recovery relies on retrying the active stage, which is insufficient when a failure irreversibly alters the scene (e.g., a dropped object rolling away); future work should explore state-aware dynamic replanning. Finally, our dependency verification is limited to adjacent subtasks, acting as a short-term memory. Extending the manager with long-term episodic memory to track and enforce non-adjacent dependencies across distant stages is a critical next step for truly long-horizon autonomy.
\clearpage
\twocolumn
\raggedbottom
\balance
\bibliographystyle{plainnat}
\bibliography{references}

\end{document}